\documentclass{article}

    \usepackage[preprint]{neurips_2024}

\usepackage[utf8]{inputenc} % allow utf-8 input
\usepackage[T1]{fontenc}    % use 8-bit T1 fonts
\usepackage{hyperref}       % hyperlinks
\usepackage{url}            % simple URL typesetting
\usepackage{booktabs}       % professional-quality tables
\usepackage{amsfonts}       % blackboard math symbols
\usepackage{nicefrac}       % compact symbols for 1/2, etc.
\usepackage{microtype}      % microtypography
\usepackage{xcolor}         % colors
\usepackage{graphicx}
\usepackage{adjustbox}
\usepackage{rotating}
\usepackage{amsmath}
\usepackage{multirow}
\usepackage{listings}
\usepackage{xcolor}

\definecolor{codegreen}{rgb}{0,0.6,0}
\definecolor{codegray}{rgb}{0.5,0.5,0.5}
\definecolor{codepurple}{rgb}{0.58,0,0.82}
\definecolor{backcolour}{rgb}{0.95,0.95,0.92}

\lstdefinestyle{mystyle}{
    backgroundcolor=\color{backcolour},   
    commentstyle=\color{codegreen},
    keywordstyle=\color{magenta},
    numberstyle=\tiny\color{codegray},
    stringstyle=\color{codepurple},
    basicstyle=\ttfamily\footnotesize,
    breakatwhitespace=false,         
    breaklines=true,                 
    captionpos=b,                    
    keepspaces=true,                 
    numbers=left,                    
    numbersep=5pt,                  
    showspaces=false,                
    showstringspaces=false,
    showtabs=false,                  
    tabsize=2
}

\title{
         FEET: A Framework for Evaluating Embedding Techniques \includegraphics[scale=0.05]{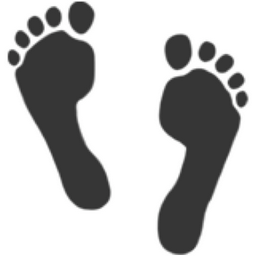}
}

\author{%
  Simon A. Lee\\
  Department of Computational Medicine\\
  UCLA\\
  \texttt{simonlee711@g.ucla.edu} \\
  \And
  John Lee \\
  Department of Computer Science \\
  UCLA\\
  \texttt{tmdgjs2592@g.ucla.edu} \\
  \AND
  Jeffrey N. Chiang \\
  Department of Computational Medicine\\
  UCLA\\
  \texttt{njchiang@g.ucla.edu} \\
}

\begin{document}

\maketitle

\begin{abstract}
In this study, we introduce FEET, a standardized protocol designed to guide the development and benchmarking of foundation models. While numerous benchmark datasets exist for evaluating these models, we propose a structured evaluation protocol across three distinct scenarios to gain a comprehensive understanding of their practical performance. We define three primary use cases: frozen embeddings, few-shot embeddings, and fully fine-tuned embeddings. Each scenario is detailed and illustrated through two case studies: one in sentiment analysis and another in the medical domain, demonstrating how these evaluations provide a thorough assessment of foundation models' effectiveness in research applications. We recommend this protocol as a standard for future research aimed at advancing representation learning models.
\end{abstract}

\section{Introduction}

Foundation models, such as BERT \citep{devlin2018bert}, GPT \citep{radford2018improving}, and CLIP \citep{radford2021learning}, have revolutionized the field of artificial intelligence. These models are trained on vast datasets using self-supervised methods \citep{ericsson2022self}, with little to no explicit supervision, enabling them to capture a wide range of knowledge. Once pre-trained, they can be fine-tuned for specific applications \citep{bommasani2021opportunities}, making them highly versatile. Their success has led to widespread adoption across various domains, including biomedicine \citep{wornow2023shaky}, astronomy \citep{lanusse2023astroclip}, and particle physics \citep{birk2024omnijet}. The rapid proliferation of foundation models has spurred continuous development, with frequent updates shared on platforms like Hugging Face \citep{wolf2019huggingface}. While existing benchmark datasets, such as MMLU \citep{hendrycks2020measuring} and GSM8k \citep{cobbe2021training}, measure the performance of these models on specific tasks, standardized protocols for evaluating and reporting their performance metrics remain underdeveloped.

To address this gap, we introduce the \textit{Framework for Evaluating Embedding Techniques (FEET)}, a comprehensive protocol that categorizes foundation model use cases into three types: frozen embeddings, few-shot embeddings, and fully fine-tuned embeddings. By systematically evaluating these categories, we aim to assess the adaptability and effectiveness of foundation models, providing a holistic view of their practical utility. We demonstrate the application of this evaluation protocol through two case studies: the first is a canonical example in natural language processing, focusing on sentiment analysis using the SST-2 (Stanford Sentiment Treebank 2) dataset \citep{socher2013recursive}. The second is a medical case study that builds on the work of Lee et al. (2024) \citep{lee2024enhancing}, highlighting the relevance and importance of this approach based on our interesting findings.

\vspace{-0.15cm}

\paragraph{Why is this an important problem?}

\begin{figure}[t]
\centering
\includegraphics[width=0.9\textwidth]{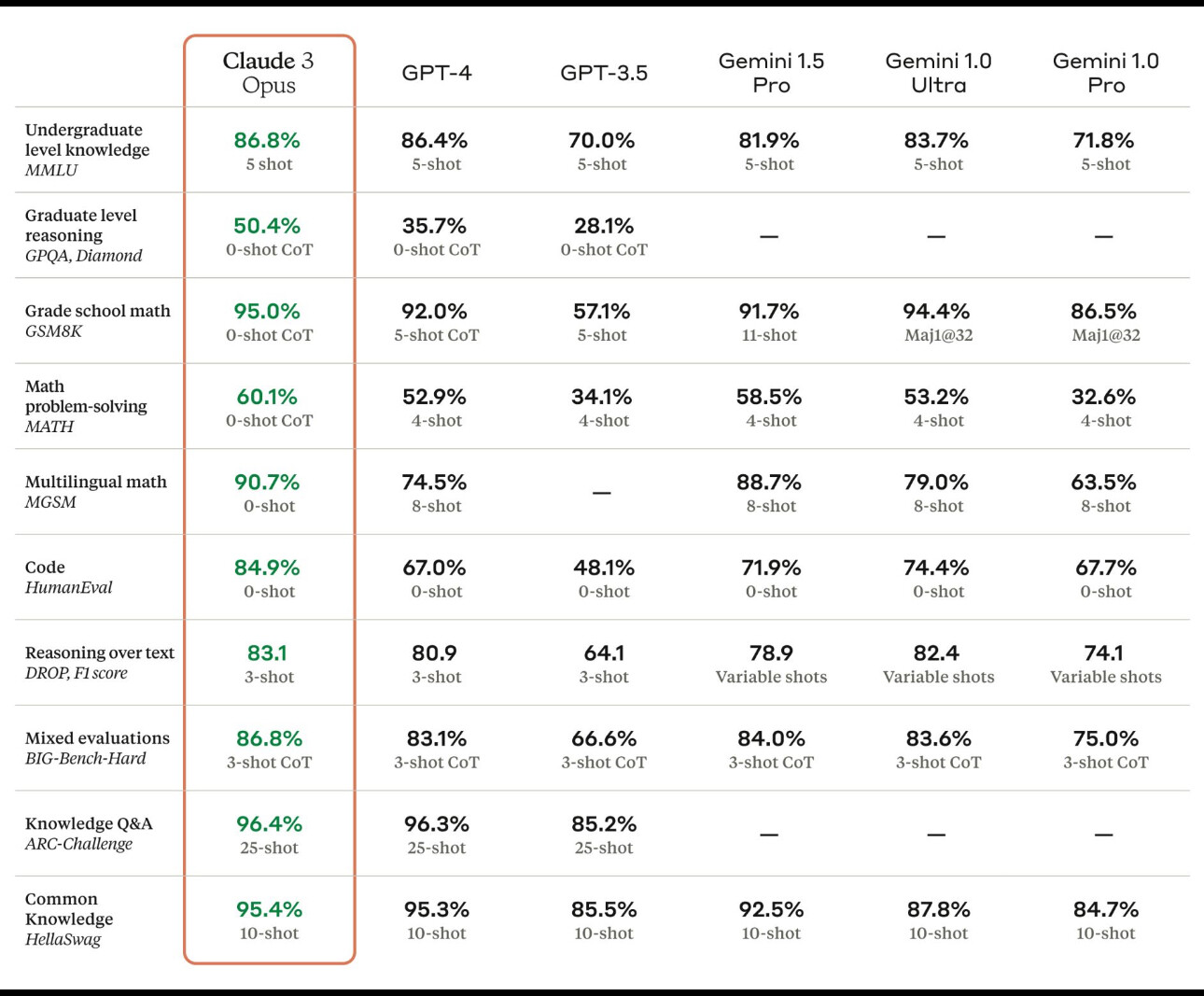}
\caption{A comparative analysis of the Claude Model, GPT, and Gemini across varying shot counts. The selection of shot numbers (0, 25, 4, 3, 10) appears arbitrary and inconsistent, raising concerns about potential cherry-picking to emphasize Claude as the state-of-the-art (SOTA) model.}
\label{fig:wild}
\end{figure}

Rigorous benchmarking protocols are crucial as they establish performance standards and promote competition, which in turn drives advancements in machine learning and more broadly in science. A well-defined protocol facilitates reproducibility and provides a standardized framework for evaluating diverse models and approaches, allowing for consistent progress from one study to the next. Currently, benchmarks in foundation model research suffer from inadequate evaluation practices, creating uncertainties about what truly constitutes state-of-the-art (SOTA). These benchmarks often rely on numerous datasets, lack reproducibility \citep{theis2015note}, and span multiple tasks, which complicates comprehensive assessments of their generalizability and potential for future improvements. We propose that this protocol will address these gaps, enabling more thorough and standardized evaluations of model capabilities within the broader scientific community.

To further motivate the problem, we examine an example in the wild (Figure \ref{fig:wild}), which compares Large Language Models (LLMs) on various benchmarking datasets. However, several key issues arise from this comparison. When the LLMs were evaluated on Grade School Math (GSM8K), the number of shots measured across models was inconsistent. There were subtle variations between 0-shot Chain of Thought (CoT) and 5-shot CoT, making it difficult to draw accurate comparisons. Another notable issue is the misalignment of shot counts across different datasets. Some datasets present state-of-the-art performance at zero shots, while others go as high as 25 shots. These variable shot numbers are not well-motivated, and we argue that such inconsistent reporting practices are detrimental to scientific progress. In a field already overwhelmed by the sheer volume of AI research, we urge researchers to avoid cherry-picking examples and to adopt standardized, transparent evaluation methodologies.

\section{Background}

Foundation model embeddings can be categorized into three primary use cases: Frozen Embeddings, Few-shot Embeddings, and Fine-tuned Embeddings. While many studies introducing new foundation models explore different permutations of these embeddings, we argue that it is crucial to present all three types for a comprehensive evaluation. Doing so ensures a more thorough assessment of a model's performance and is necessary to substantiate claims of achieving state-of-the-art results.

\subsection{Frozen Embeddings}

Frozen embeddings refer to pre-trained feature embeddings that remain unchanged during the training of a new model. The term `frozen' indicates that these embeddings retain their pre-trained parameter values without any further modification: \( \mathbf{E}_{\text{frozen}} = \mathbf{E}_{\text{initial}} \). Prior to the rise of transformer-based foundation models, embedding techniques such as Word2Vec \citep{church2017word2vec}, GloVe \citep{pennington2014glove}, and fastText \citep{joulin2016fasttext} were widely used as frozen embeddings in various applications.

Frozen embeddings are particularly valuable in machine learning, especially in scenarios where the generality and robustness of pre-trained models are crucial. Studies such as \citep{lee2024multimodal} demonstrate how frozen embeddings from open-source models can be applied across a wide range of clinical tasks without requiring any fine-tuning. By leveraging the extensive knowledge embedded in models trained on large, diverse datasets, researchers and practitioners can evaluate how well these pre-trained representations transfer to new tasks or domains without the need for additional training.

\subsection{Few-Shot Embeddings}

Few-shot learning is a machine learning technique that enables models to learn from a limited number of labeled examples, which is particularly important in scenarios where extensive data collection is impractical, such as specialized fields or rare event detection \citep{brown2020language, parnami2022learning, lai2020extensively}. It operates by distinguishing between a `support set' \( S = \{(x_i, y_i)\}_{i=1}^k \), which contains a few labeled examples for training, and a `query set' \( Q = \{x'_j\}_{j=1}^m \), which consists of samples for classification. The goal is to train a model, represented by \( f \) and parameterized by \( \theta \), to predict the labels of the query set by optimizing a similarity function:

\[ y'_j = \arg\max_y \text{sim}(f_\theta(x'_j), f_\theta(x_i)) \quad \forall (x_i, y_i) \in S \]

Foundation models significantly enhance the effectiveness of few-shot learning by leveraging their deep architectures and self-attention mechanisms \citep{vaswani2017attention} to transfer learned features with minimal data \citep{wang2020generalizing}. These transferable embeddings closely align with the concept of transfer learning \citep{weiss2016survey, torrey2010transfer}. However, the current literature on few-shot learning often highlights inconsistencies in evaluation practices, such as variable shot numbers and arbitrary shot selection across different studies as described in Figure \ref{fig:wild}. 

\subsection{Fine-tuned Embeddings}

Fine-tuning is a machine learning technique that adapts a foundation model to perform effectively on specific datasets tailored for particular tasks or domains \citep{dodge2020fine}. This process involves optimizing the model's parameters \( \Theta \) to minimize a loss function \( \mathcal{L} \) on the dataset \( \mathcal{D} \):

\[
\Theta^* = \arg \min_{\Theta} \mathcal{L}(\Theta; \mathcal{D})
\]

where \( \Theta^* \) represents the parameters optimized for the task. A notable example of fine-tuning is LoRA (Low-Rank Adaptation) \citep{hu2021lora}, which modifies the weight matrix \( \mathbf{W}_l \) of a layer \( l \) by incorporating the product of two low-rank matrices, \( \mathbf{A}_l \) and \( \mathbf{B}_l \):

\[
\mathbf{W}_l' = \mathbf{W}_l + \mathbf{A}_l \mathbf{B}_l
\]

This approach retains the original structure of the model while efficiently adapting it to new tasks, reducing the number of trainable parameters. Other techniques, such as adapter layers \citep{sung2022vl, karimi2021compacter} and Parameter Efficient Fine-Tuning (PEFT) \citep{ding2023parameter}, also allow for effective model customization with minimal structural changes. These strategies enable the application of pre-trained models to new tasks, leveraging existing knowledge while enhancing performance on specialized tasks. 

Many studies have used finetuning as a basic technique to use foundations models for specific tasks \citep{ono2024text, jiang2024supervised}. A key insight is that fine-tuning with a large sample size is generally preferred, as it reduces the risk of overfitting and improves the model's ability to generalize to new data, ultimately optimizing performance across a broader range of tasks \citep{majdik2024sample}.

\section{Methods}

\subsection{A Framework for Evaluating Embedding Techniques (FEET)}

In this section we introduce a Framework for Evaluating Embedding Techniques (FEET). We propose this protocol because a comprehensive evaluation of foundation models provides details under different experimental conditions and the adaptability of a foundation model to a given task. We described the three key embedding types in our previous section (Frozen Embeddings, Few-shot Embeddings, and Fine-tuned Embeddings) and encourage that these embeddings are reported. This structured approach is essential for understanding each model's adaptability and performance across different levels of customization.

Frozen Embeddings assess the model's capabilities as they emerge directly from pre-training, providing valuable insights into its out-of-the-box utility. Few-shot Embeddings gauge the model’s ability to adapt to new tasks with minimal examples, highlighting its efficiency in rapid learning. We recommend standardizing the evaluation of few-shot learning by reporting performance at powers of \(2^N\), up to \(2^{10}\), to ensure consistency across studies. Finally, Fine-tuned Embeddings reflect the model’s peak performance after extensive training on specific datasets (assuming sufficient sample size), offering a view of its optimal capabilities for specialized tasks.

\subsubsection{Measuring Deltas \(\Delta\)}

Evaluating foundation models through Frozen, Few-shot, and Fine-tuned embeddings provides essential insights into their performance, with \(\Delta\) quantifying improvements or potential drop-offs—critical for guiding model development. This allows for the identification of models that offer the most significant enhancements, making \(\Delta\) a valuable metric for model selection.

\(\Delta\) enables comparative analysis across embedding types, illuminating the trade-offs between ease of implementation and performance gains. This is crucial for choosing the right model and technique for specific tasks. For instance, tasks requiring high accuracy may justify the resource-intensive fine-tuning process, whereas tasks needing rapid deployment may benefit from less demanding Frozen or Few-shot embeddings.

Mathematically, \(\Delta\) represents the performance differential between embeddings, providing numerical insights into improvements for each type:
\[
\Delta_{\text{Few-shot}_{i}} = \text{Performance}_{\text{Few-shot}_{i}} - \text{Performance}_{\text{Frozen}}
\]
\[
\Delta_{\text{Fine-tuned}} = \text{Performance}_{\text{Fine-tuned}} - \text{Performance}_{\text{Frozen}}
\]
These metrics offer a clear measure of performance improvement, supporting a more comprehensive evaluation across various dimensions. Applying statistical tests to \(\Delta\) values further validates the significance of observed differences, enhancing the rigor of the evaluation.

\subsubsection{FEET Tables}

The FEET Table offers a structured and comprehensive way to present the performance of embeddings across three distinct use cases: Frozen, Few-shot, and Fine-tuned. Each row corresponds to a different model, detailing specific metrics that reflect how each model performs under these varying conditions. This table helps provide a complete view of the model’s versatility and effectiveness across different levels of customization. An example is shown in Table 1.

\begin{table}[ht]
\centering
\caption{An Example of a FEET Table. This table provides a detailed comparison of embedding performance across the Frozen, Few-shot, and Fine-tuned scenarios.}
\begin{tabular}{llll}
\textbf{Models} & \textbf{Frozen} & \textbf{Few-shot} & \textbf{Fine-tuned} \\
Model A & X.XXX & Y.YYY & Z.ZZZ \\
Model B & X.XXX & Y.YYY & Z.ZZZ \\
\end{tabular}
\label{tab:FEET_Table}
\end{table}

Additionally, the \(\Delta\) FEET Table focuses on illustrating the relative improvements or declines in performance, using the Frozen embedding as a baseline to compute \(\Delta\) for the Few-shot and Fine-tuned scenarios. This table is formatted to emphasize percentage changes, offering a clear visual comparison of how much each model improves or adapts when additional training or data specificity is applied. Such comparisons are invaluable for determining the trade-offs between minimal adjustments (Few-shot) and more resource-intensive approaches (Fine-tuning). An example is shown in Table 2.

\begin{table}[ht]
\centering
\caption{An Example of a \(\Delta\) FEET Table. This table highlights the relative performance changes (percentage improvements) between Frozen, Few-shot, and Fine-tuned embeddings.}
\begin{tabular}{llll}
\textbf{Models} & \textbf{Frozen} & \textbf{Few-shot} & \textbf{Fine-tuned} \\
Model A & X.XXX & Y.YYY (\(\Delta\%\)) & Z.ZZZ (\(\Delta\%\)) \\
Model B & X.XXX & Y.YYY (\(\Delta\%\)) & Z.ZZZ (\(\Delta\%\)) \\
\end{tabular}
\label{tab:delta_FEET_Table}
\end{table}

\subsection{Benchmarking Experiments}

\subsubsection{Case Study: Sentiment Analysis}

In our first case study, we provide a canonical example in the field of Natural Language Processing where we evaluate three foundation models—BERT (2018) \citep{devlin2018bert}, DistilBERT (2020) \citep{sanh2019distilbert}, and GPT-2 (2019) \citep{radford2019language}—on the task of sentiment analysis using the SST-2 (Stanford Sentiment Treebank 2) dataset \citep{socher2013recursive}. Each model is fine-tuned to perform binary classification, where the goal is to classify movie reviews as either positive or negative. This evaluation was conducted to provide a quick, demonstrative comparison between transformer-based models (encoder-based and decoder-based), and their use in this study highlights simplicity and speed of deployment rather than optimization for the task. The dataset consists of thousands of sentences, and we applied a 70/30 split for training, and testing.

For evaluation, we focus on four metrics for this case study: accuracy, precision, recall, and F1 score. Accuracy measures the overall correctness of the model, while precision and recall offer a deeper understanding of how well the models handle false positives and false negatives. The F1 score balances these two factors. These metrics offer a flexible framework for testing models in quick, real-world scenarios, showcasing how easily sentiment classification tasks can be adapted using general-purpose language models.

\subsubsection{Case Study: Antibiotic Susceptibility Prediction}

In the second case study, we evaluate three biomedical foundation models—Bio\_ClinicalBERT (2019) \citep{alsentzer2019publicly}, MedBERT (2022) \citep{vasantharajan2022medbert}, and SciBERT (2019) \citep{beltagy2019scibert}—on the task of predicting the susceptibility of patient for recieving multiple antibiotics (Clindamycin, Erythromycin, Gentamicin, Levofloxacin, Oxacillin, Tetracycline, Trimethoprim, and Vancomycin), as detailed in \citep{lee2024enhancing}. Each model is equipped with a linear prediction head to perform binary classification for each antibiotic. The task is to identify whether a patient based on their previous medical history is ``susceptible'' or ``not susceptible''. We use the MIMIC-IV ED dataset \citep{johnson2020mimic}, and define an inclusion criteria consisting of 5,976 patient records. We apply a 70/30 split for training, validation, and testing. 

For evaluation, we focus on two new metrics: Area Under the Receiver Operating Characteristic Curve (AUROC) and Area Under the Precision-Recall Curve (AUPRC). We use two new metrics in this case study to show the flexibility of our framework. AUROC is used to assess model performance in distinguishing between positive and negative cases, while AUPRC is particularly valuable for imbalanced datasets, where precision and recall are critical. These metrics offer a comprehensive view of each model’s predictive capabilities, especially for tasks like antibiotic prediction, where both false positives and false negatives can have serious clinical implications. We note that these performance metrics can be swapped with other metrics based on the domain and foundation model being used (classification vs. generation-based metrics).

\subsubsection{Foundation Model Hyperparameters}

We implement all stages utilizing the Huggingface transformers framework\footnote{\href{https://huggingface.co/docs/transformers/en/index}{https://huggingface.co/docs/transformers/en/index}}. The default AdamW optimizer, along with a linear learning rate scheduler, is used throughout all training phases. All foundation models are evaluated for 100 epochs in the fine-tuning with early stopping implemented at a learning rate of $2 \times 10^{-5}$, after which the model approaches convergence. During the finetuning phase, checkpoints are created every 50 minibatches. The model is then restored from the checkpoint exhibiting the lowest validation loss, monitored across a maximum of 10 epochs. The entire computational pipeline is executable on a single 48GB A100 GPU well under a 24-hour timeframe.

\section{Results}

\subsection{Case Study: Sentiment Analysis}

\begin{table}[t!]
\centering
\caption{Accuracy, Precision, Recall, and F1 Scores (\%) with 95\% Confidence Intervals for Different Models and Sample Sizes}
\label{tab:metrics}
\begin{adjustbox}{width=1\textwidth}
\begin{tabular}{llcccccccccccc}
\toprule
\textbf{Metric} & \textbf{Models} & \textbf{Frozen} & \textbf{2-shot} & \textbf{4-shot} & \textbf{8-shot} & \textbf{16-shot} & \textbf{32-shot} & \textbf{64-shot} & \textbf{128-shot} & \textbf{256-shot} & \textbf{512-shot} & \textbf{1024-shot} & \textbf{Fine-Tuned} \\
\midrule

\multirow{3}{*}{\textbf{Accuracy}} & \textbf{BERT} & 47.50 (0.01) & 52.75 (0.14) & 51.25 (0.21) & \textbf{53.50} (0.18) & 49.50 (0.11) & \textbf{52.50} (0.27) & \textbf{59.00} (0.23) & \textbf{78.50} (0.19) & \textbf{85.75} (0.09) & 80.50 (0.14) & 82.00 (0.12) & \textbf{86.00} (0.13) \\
& \textbf{DistilBERT} & \textbf{47.75} (0.02) & \textbf{53.75} (0.17) & \textbf{52.25} (0.30) & 52.50 (0.28) & \textbf{52.50} (0.28) & \textbf{52.50} (0.29) & 52.50 (0.28) & 65.50 (0.20) & 83.00 (0.07) & \textbf{80.75} (0.02) & \textbf{85.25} (0.12) & 85.50 (0.10) \\
& \textbf{GPT2} & \textbf{47.75} (0.04) & 53.00 (0.28) & 49.50 (0.23) & 49.50 (0.15) & 48.25 (0.29) & 49.25 (0.11) & 48.75 (0.17) & 50.50 (0.16) & 53.00 (0.11) & 55.50 (0.05) & 75.75 (0.11) & 80.00 (0.04) \\
\midrule

\multirow{3}{*}{\textbf{Precision}} & \textbf{BERT} & 48.79 (0.56) & 53.44 (0.32) & \textbf{48.34} (0.30) & \textbf{53.06} (0.27) & \textbf{49.63} (0.24) & 27.56 (0.35) & \textbf{65.80} (0.32) & \textbf{81.26} (0.22) & \textbf{86.08} (0.14) & 81.15 (0.18) & 84.97 (0.15) & \textbf{87.24} (0.16) \\
& \textbf{DistilBERT} & 57.61 (0.26) & 53.38 (0.28) & 27.50 (0.37) & 27.56 (0.37) & 27.56 (0.37) & 27.56 (0.37) & 27.56 (0.36) & 69.31 (0.31) & 83.67 (0.15) & \textbf{81.56} (0.08) & \textbf{85.25} (0.19) & 86.02 (0.15) \\
& \textbf{GPT2} & \textbf{75.12} (2.95) & \textbf{53.69} (0.53) & 47.51 (0.35) & 50.89 (0.35) & 47.48 (0.41) & \textbf{50.35} (0.34) & 47.47 (0.30) & 52.81 (0.46) & 53.12 (0.26) & 55.55 (0.19) & 76.01 (0.20) & 80.33 (0.13) \\
\midrule

\multirow{3}{*}{\textbf{Recall}} & \textbf{BERT} & 47.50 (0.01) & 52.75 (0.14) & 51.25 (0.21) & \textbf{53.50} (0.19) & 49.50 (0.10) & \textbf{52.50} (0.28) & \textbf{59.00} (0.24) & \textbf{78.50} (0.18) & \textbf{85.75} (0.09) & 80.50 (0.14) & 82.00 (0.12) & \textbf{86.00} (0.13) \\
& \textbf{DistilBERT} & \textbf{47.75} (0.02) & \textbf{53.75} (0.17) & \textbf{52.25} (0.30) & 52.50 (0.28) & \textbf{52.50} (0.28) & \textbf{52.50} (0.28) & 52.50 (0.28) & 65.50 (0.20) & 83.00 (0.07) & \textbf{80.75} (0.03) & \textbf{85.25} (0.12) & 85.50 (0.10) \\
& \textbf{GPT2} & \textbf{47.75} (0.03) & 53.00 (0.27) & 49.50 (0.24) & 49.50 (0.14) & 48.25 (0.29) & 49.25 (0.11) & 48.75 (0.17) & 50.50 (0.17) & 53.00 (0.12) & 55.50 (0.05) & 75.75 (0.11) & 80.00 (0.04) \\
\midrule

\multirow{3}{*}{\textbf{F1 Score}} & \textbf{BERT} & \textbf{31.88} (0.02) & \textbf{52.42} (0.14) & 42.19 (0.25) & \textbf{50.25} (0.22) & \textbf{49.53} (0.11) & 36.15 (0.34) & \textbf{52.17} (0.25) & \textbf{77.79} (0.17) & \textbf{85.67} (0.10) & 80.29 (0.15) & 81.44 (0.10) & \textbf{85.80} (0.13) \\
& \textbf{DistilBERT} & 31.57 (0.04) & 50.59 (0.19) & 36.03 (0.33) & 36.15 (0.34) & 36.15 (0.34) & 36.15 (0.36) & 36.15 (0.35) & 62.88 (0.19) & 82.98 (0.08) & \textbf{80.51} (0.02) & \textbf{85.25} (0.12) & 85.39 (0.10) \\
& \textbf{GPT2} & 31.14 (0.07) & 41.09 (0.34) & \textbf{45.34} (0.24) & 46.99 (0.19) & 47.20 (0.32) & \textbf{47.45} (0.11) & 46.57 (0.22) & 46.69 (0.17) & 53.03 (0.12) & 55.52 (0.06) & 75.58 (0.10) & 80.00 (0.04) \\
\bottomrule

\end{tabular}
\end{adjustbox}
\end{table}

\begin{table}[t!]
\centering
\caption{Corresponding $\Delta$ FEET Table describing the Accuracy, Precision, Recall and F1 Score Differentials on the SST-2 Sentiment Analysis Task.}
\label{tab:delta_metrics}
\begin{adjustbox}{width=1\textwidth}
\begin{tabular}{llcccccccccccc}
\toprule
\textbf{Metric} & \textbf{Models} & \textbf{Frozen} & \textbf{2-shot} & \textbf{4-shot} & \textbf{8-shot} & \textbf{16-shot} & \textbf{32-shot} & \textbf{64-shot} & \textbf{128-shot} & \textbf{256-shot} & \textbf{512-shot} & \textbf{1024-shot} & \textbf{Fine-Tuned} \\
\midrule

\multirow{3}{*}{\textbf{Accuracy}} & \textbf{BERT} & ------ & 5.25\% & 3.75\% & 6.00\% & 2.00\% & 5.00\% & 11.50\% & 31.00\% & 38.25\% & 33.00\% & 34.50\% & 38.50\% \\
& \textbf{DistilBERT} & ------ & 6.00\% & 4.50\% & 4.75\% & 4.75\% & 4.75\% & 4.75\% & 17.75\% & 35.25\% & 33.00\% & 37.50\% & 37.75\% \\
& \textbf{GPT2} & ------& 5.25\% & 1.75\% & 1.75\% & 0.50\% & 1.50\% & 1.00\% & 2.75\% & 5.25\% & 7.75\% & 28.00\% & 32.25\% \\
\midrule

\multirow{3}{*}{\textbf{Precision}} & \textbf{BERT} & ------ & 4.65\% & -0.45\% & 4.27\% & 0.84\% & -21.23\% & 17.01\% & 32.47\% & 37.29\% & 32.36\% & 36.18\% & 38.45\% \\
& \textbf{DistilBERT} & ------ & -4.23\% & -30.11\% & -30.05\% & -30.05\% & -30.05\% & -30.05\% & 11.70\% & 26.06\% & 23.95\% & 27.64\% & 28.41\% \\
& \textbf{GPT2} & ------ & -21.43\% & -27.61\% & -24.23\% & -27.64\% & -24.77\% & -27.65\% & -22.31\% & -22.00\% & -19.57\% & 0.89\% & 5.21\% \\
\midrule

\multirow{3}{*}{\textbf{Recall}} & \textbf{BERT} & ------ & 5.25\% & 3.75\% & 6.00\% & 2.00\% & 5.00\% & 11.50\% & 31.00\% & 38.25\% & 33.00\% & 34.50\% & 38.50\% \\
& \textbf{DistilBERT} & ------ & 6.00\% & 4.50\% & 4.75\% & 4.75\% & 4.75\% & 4.75\% & 17.75\% & 35.25\% & 33.00\% & 37.50\% & 37.75\% \\
& \textbf{GPT2} & ------ & 5.25\% & 1.75\% & 1.75\% & 0.50\% & 1.50\% & 1.00\% & 2.75\% & 5.25\% & 7.75\% & 28.00\% & 32.25\% \\
\midrule

\multirow{3}{*}{\textbf{F1 Score}} & \textbf{BERT} & ------ & 20.54\% & 10.31\% & 18.37\% & 17.65\% & 4.27\% & 20.29\% & 45.91\% & 53.79\% & 48.41\% & 49.56\% & 53.92\% \\
& \textbf{DistilBERT} & ------ & 19.02\% & 4.46\% & 4.58\% & 4.58\% & 4.58\% & 4.58\% & 31.31\% & 51.41\% & 48.94\% & 53.68\% & 53.82\% \\
& \textbf{GPT2} & ------ & 9.95\% & 14.20\% & 15.85\% & 16.06\% & 16.31\% & 15.43\% & 15.55\% & 21.89\% & 24.38\% & 44.44\% & 48.86\% \\
\bottomrule

\end{tabular}
\end{adjustbox}
\end{table}

The Feet and Delta Feet Tables in 3,4 demonstrate a clear trend of model performance improvement as the training progresses from few-shot learning to fine-tuning. Initially, with few-shot settings, models like BERT, DistilBERT, and GPT2 show lower performance across accuracy, precision, recall, and F1 scores. However, as the number of training samples increases, the performance gradually improves, with the highest scores observed in the fine-tuned models. BERT and DistilBERT, in particular, exhibit substantial gains in accuracy and recall as more data is incorporated. Precision shows more variability but still follows an upward trend. GPT2 shows smaller gains in few-shot scenarios but improves significantly with full fine-tuning. Overall, the tables confirm the ``expected'' behavior that more extensive training, especially through fine-tuning, results in better model performance across all metrics.

\subsection{Case Study: Antibiotic Susceptibility Prediction}

\begin{table}[t!]
\centering
\caption{FEET Table describing the AUROC Scores on all the tasks for various use cases of FM}
\begin{adjustbox}{width=1\textwidth}
\label{tab:FEET Table}
\begin{tabular}{llcccccccccccc}
\toprule
\textbf{AUROC/Task} & \textbf{Models} & \textbf{Frozen} & \textbf{2 shot} & \textbf{4 shot} & \textbf{8 shot} & \textbf{16 shot} & \textbf{32 shot} & \textbf{64 shot} & \textbf{128 shot} & \textbf{256 shot} & \textbf{512 shot} & \textbf{1024 shot} & \textbf{Fine-tuned} \\
\midrule
\multirow{3}{*}{\textbf{Clindamycin}} & BioClinicalBERT & \textbf{74.99 (3.86)} & 53.88 (2.89) & 54.69 (2.99) & \textbf{56.73 (2.97)} & 53.87 (2.88) & \textbf{56.67 (2.79)} & 56.84 (2.96) & 56.88 (3.01) & 57.12 (3.13) & 58.72 (2.88) & 59.55 (3.00) & 67.59 (2.65) \\
&MedBERT & 74.22 (3.74) & \textbf{54.28 (3.67)} & \textbf{55.49 (2.85)} & 55.39 (2.47) & \textbf{54.35 (3.63)} & 56.24 (2.59) & \textbf{57.27 (3.49)} & \textbf{57.56 (2.97)} & \textbf{58.32 (2.78)} & \textbf{59.28 (3.23)} & \textbf{60.86 (2.97)} & \textbf{69.35 (2.99)}  \\
&SciBERT & 73.98 (3.21) & 54.18 (3.05) & 51.60 (2.86) & 52.77 (2.94) & 50.11 (2.67) & 52.18 (3.26) & 54.10 (3.20) & 56.41 (2.88) & 56.92 (3.17) & 56.60 (3.11) & 58.88 (3.21) & 68.31 (2.88) \\
\midrule
\multirow{3}{*}{\textbf{Erythromycin}} & BioClinicalBERT & 75.38 (3.52) & 57.91 (3.15) & \textbf{58.71 (3.25)} & \textbf{61.15 (3.20)} & \textbf{60.59 (2.54)} & \textbf{58.23 (3.01)} & \textbf{60.70 (3.26)} & \textbf{60.99 (2.73)} & 60.03 (3.98) & 60.56 (3.06) & 61.78 (2.77) & 70.12 (3.20) \\
&MedBERT & 75.38 (3.45) & \textbf{59.25 (3.76)} & 56.47 (3.12) & 59.31 (2.99) & 59.53 (2.57) & 55.79 (2.53) & 55.80 (3.22) & 60.02 (3.69) & \textbf{61.51 (3.87)} & \textbf{62.57} (3.24) & \textbf{63.57 (3.12)} & \textbf{71.65 (3.01)} \\
&SciBERT & 74.27 (3.14) & 54.54 (3.92) & 52.59 (3.20) & 56.57 (2.72) & 53.84 (2.50) & 55.97 (3.33) & 57.13 (3.75) & 59.19 (3.57) & 59.69 (2.91) & 59.91 (3.15) & 59.21 (2.85) & 69.00 (2.99) \\
\midrule

\multirow{3}{*}{\textbf{Gentamicin}} & BioClinicalBERT & 68.35 (3.86) & 52.77 (2.89) & 50.79 (3.02) & \textbf{52.89 (2.97)} & \textbf{52.09 (2.88)} & 47.04 (2.79) & \textbf{50.70 (3.12)} & 49.48 (3.01) & 48.51 (3.23) & 52.61 (2.88) & 59.56 (3.00) & 63.95 (3.24) \\
&MedBERT & \textbf{73.08 (3.99)} &\textbf{ 58.65 (2.67)} & \textbf{59.91 (2.45)} & 39.83 (2.47) & 41.33 (2.63) & \textbf{53.01 (2.59)} & 37.36 (2.49) & 39.85 (2.47) & \textbf{49.23 (2.78)} & \textbf{55.57 (2.43)} & \textbf{59.98 (2.57)} &\textbf{ 67.05 (2.99)} \\
&SciBERT & 70.41 (2.95) & 41.96 (3.92) & 48.37 (3.20) & 44.67 (2.72) & 45.18 (2.50) & 47.78 (3.33) & 45.13 (3.75) & \textbf{49.61 (3.57)} & 48.85 (2.91) & 41.63 (3.15) & 47.89 (2.85) & 54.66 (2.95) \\
\midrule

\multirow{3}{*}{\textbf{Levofloxacin}} & BioClinicalBERT & 80.05 (3.23) & 59.24 (2.29) & 59.36 (3.83) & \textbf{60.90 (2.37)} & \textbf{57.48 (2.58) }& \textbf{55.33 (2.77)} & 58.81 (3.27) & 63.50 (3.11) & 65.27 (3.20) & 66.46 (2.78) & 67.26 (3.33) & 69.20 (3.20) \\
&MedBERT & 77.94 (3.79) & \textbf{60.06 (2.37)} & \textbf{60.46 (2.44)} & 55.69 (2.41) & 51.99 (2.62) & 54.89 (2.55) &\textbf{ 60.52 (2.46)} & \textbf{64.75 (2.27)} & \textbf{66.05 (2.72}) & \textbf{66.87 (2.41)} & 67.92 \textbf{(2.51)} & \textbf{69.50 (2.17)} \\
&SciBERT & \textbf{80.30 (2.93)} & 54.06 (3.91) & 52.19 (3.24) & 52.62 (2.72) & 54.26 (2.52) & 43.30 (3.13) & 58.50 (3.15) & 63.50 (3.37) & 64.31 (2.81) & 65.06 (3.85) & 66.45 (2.88) & 68.17 (2.92) \\
\midrule

\multirow{3}{*}{\textbf{Oxacillin}} & BioClinicalBERT & 77.85 (3.16) & 36.12 (2.49) & 35.37 (3.22) & 49.45 (2.27) & \textbf{62.43 (2.58)} & \textbf{58.86 (2.59)} & \textbf{63.86 (3.32)} & 62.77 (3.31) & \textbf{64.91 (3.03)} & \textbf{64.82 (2.78)} & \textbf{65.84 (3.40)} & 66.91 (3.34) \\
&MedBERT & 77.09 (3.39) & 37.98 (2.37) & \textbf{40.85 (2.25)} & \textbf{54.61 (2.87)} & 56.78 (2.11) & 54.78 (2.52) & 59.31 (2.49) & \textbf{63.32 (2.49)} & 63.34 (2.77) & 63.78 (2.41) & 64.61 (2.28) & 65.70 (2.50) \\
&SciBERT & \textbf{78.47 (2.23)} & \textbf{41.65 (3.33)} & 38.51 (3.27) & 50.27 (2.44) & 53.35 (2.52) & 49.04 (3.29) & 59.96 (3.25) & 62.78 (3.77) & 62.97 (2.71) & 64.50 (3.45) & 65.26 (2.44) & \textbf{67.11 (2.44)} \\

\midrule
\multirow{3}{*}{\textbf{Tetracycline}} & BioClinicalBERT & 67.28 (3.32) & 52.98 (2.49) & 51.15 (3.35) & 48.19 (2.89) & 47.64 (2.81) & 51.11 (2.67) & 50.34 (3.71) & 47.50 (3.11) & 51.17 (3.44) & \textbf{53.35 (2.84)} & 52.44 (3.11) & 54.69 (3.77) \\
&MedBERT & 64.58 (3.98) & \textbf{54.84 (2.44)} & \textbf{51.99 (2.42)} & 44.98 (2.46) & \textbf{49.86 (2.43)} & 45.06 (2.79) & 45.98 (2.77) & 47.02 (2.97) & 50.56 (2.70) & 52.48 (2.63) & \textbf{56.63 (2.99)} & \textbf{57.51 (2.99)} \\
&SciBERT & \textbf{67.55 (2.55)} & 51.21 (3.95) & 50.76 (3.33) & \textbf{49.29 (2.22)} & 47.25 (2.58) & \textbf{52.09 (3.33)} & \textbf{54.36 (3.75)} & \textbf{54.21 (3.41)} &\textbf{ 54.80 (2.67)} & 53.05 (3.51) & 54.54 (2.33) & 55.55 (2.15) \\
\midrule

\multirow{3}{*}{\textbf{Trimethoprim/sulfa}} & BioClinicalBERT & 70.60 (3.16) & 56.02 (2.81) & \textbf{55.28 (3.44)} & \textbf{49.95 (2.45)} & 47.52 (2.48) & 50.24 (2.99) & \textbf{49.11 (3.32)} & 47.67 (3.71) & 53.05 (3.27) & 56.65 (2.87) & \textbf{60.17 (3.01)} & \textbf{62.06 (3.11)} \\
&MedBERT & 72.28 (3.19) & \textbf{56.22 (2.77)} & 55.11 (2.49) & 46.33 (2.67) & 46.56 (2.66) & \textbf{54.65 (2.29)} & 48.79 (2.45) & \textbf{52.17 (2.37)} & 54.39 (2.73) & 57.21 (2.13) & 59.80 (2.77) & 61.78 (2.86) \\
&SciBERT & \textbf{75.08 (2.93)} & 47.45 (3.54) & 48.48 (3.56) & 49.30 (2.98) & \textbf{48.11 (2.51)} & 51.52 (3.49) & 42.59 (3.50) & 47.73 (3.33) & \textbf{55.35 (2.78)} & \textbf{58.20 (3.16)} & 58.83 (2.56) & 60.41 (2.11) \\
\midrule
\multirow{3}{*}{\textbf{Vancomycin}} & BioClinicalBERT & 77.14 (3.81) & 41.02 (2.29) & 40.07 (3.22) & \textbf{53.45 (2.93)} & \textbf{60.07 (2.84)} & 47.41 (2.33) & 49.01 (3.68) & \textbf{62.64 (3.56)} & \textbf{63.13 (3.39)} & 63.43 (2.85) & 63.84 (3.55) & 65.77 (3.00) \\
&MedBERT & 75.93 (3.44) & 37.67 (2.55) & 37.74 (2.75) & 44.60 (2.65) & 45.10 (2.62) & 43.87 (2.59) & 45.90 (2.61) & 57.79 (2.77) & 61.21 (2.79) & 63.78 (2.61) & \textbf{64.58 (2.58)} & 66.10 (2.88) \\
&SciBERT & \textbf{78.28 (2.91)} & \textbf{47.70 (3.97)} & \textbf{42.41 (3.22)} & 51.46 (2.72) & 55.02 (2.50) & \textbf{52.39 (3.13)} & \textbf{55.85 (3.74)} & 57.94 (3.97) & 61.16 (2.21) & \textbf{64.41 (3.17)} & 64.18 (2.44) & \textbf{66.52 (2.77)} \\
\bottomrule

\toprule
\textbf{AUPRC/Task} &  &  &  &  &  &  &  &  &  & &  &  & \\
\midrule
\multirow{3}{*}{\textbf{Clindamycin}} 
&BioClinicalBERT &  78.75 (4.70) & 49.12 (4.85) & 47.48 (4.67) & 50.33 (4.82) & 53.42 (4.73) & 55.26 (4.85) & 57.64 (4.89) & 60.38 (4.77) & 61.72 (4.91) & 63.85 (4.81) & 65.94 (4.74) & 70.01 (4.90)\\
&MedBERT & 77.34 (4.54) & 48.45 (4.70) & 46.87 (4.62) & 49.71 (4.80) & 52.19 (4.67) & 54.05 (4.75) & 56.34 (4.85) & 59.22 (4.77) & 61.38 (4.89) & 63.19 (4.79) & 65.44 (4.70) & 69.45 (4.83)\\
&SciBERT & 78.02 (4.82) & 49.34 (4.78) & 47.65 (4.69) & 50.52 (4.86) & 53.14 (4.75) & 55.00 (4.84) & 57.45 (4.89) & 60.12 (4.79) & 61.95 (4.88) & 63.67 (4.82) & 66.02 (4.73) & 70.25 (4.91)\\
\midrule
\multirow{3}{*}{\textbf{Erythromycin}} & BioClinicalBERT & 69.35 (5.92) & 47.54 (5.13) & 47.71 (4.83) & 49.39 (5.26) & 49.27 (5.41) & 49.23 (4.92) & 48.93 (5.09) & 48.93 (5.87) & 47.13 (5.29) & 47.32 (4.98) & 47.20 (5.75) & 54.08 (5.03)
 \\
&MedBERT & 69.15 (5.96) & 44.70 (5.91) & 43.81 (6.12) & 47.68 (5.87) & 47.01 (6.05) & 46.10 (5.82) & 44.74 (6.18) & 46.04 (5.97) & 47.06 (6.11) & 47.37 (5.89) & 47.21 (5.67) & 55.01 (5.88)\\
&SciBERT & 68.49 (5.69) & 44.62 (5.19)& 44.94 (5.59) & 45.25 (5.76)& 44.76 (5.33) & 45.48 (5.43) & 47.55 (5.55) & 45.63 (5.19) & 45.42 (5.69) & 46.61 (5.89) & 46.14 (5.50) & 54.89 (5.77)\\
\midrule

\multirow{3}{*}{\textbf{Gentamicin}} & BioClinicalBERT & 97.02 (1.72) & 95.61 (1.83) & 95.59 (2.08) & 95.22 (1.74) & 95.09 (1.85) & 95.27 (1.94) & 95.37 (2.11) & 95.38 (1.78) & 94.89 (1.99) & 95.05 (1.88) & 95.36 (1.98) & 96.83 (1.92) \\

&MedBERT & 97.33 (1.73) & 96.27 (1.92) & 96.50 (2.05) & 94.38 (1.89) & 95.05 (2.01) & 95.68 (1.95) & 95.68 (1.87) & 92.19 (1.98) & 93.79 (2.03) & 95.05 (1.84) & 94.73 (2.07) & 96.02 (1.90) \\

&SciBERT & 96.69 (2.18) & 94.70 (1.98) & 95.53 (1.95)& 95.22 (2.03) & 95.62 (1.77) & 95.62(1.66) & 95.57 (1.70) & 95.80 (2.11) & 94.45 (2.08)& 94.54 (2.33) & 95.16 (1.67) & 96.02 (2.15) \\
\midrule

\multirow{3}{*}{\textbf{Levofloxacin}} & BioClinicalBERT & 84.41 (4.12) & 69.57 (4.10) & 68.04 (4.02) & 68.31 (3.89) & 66.22 (4.25) & 64.87 (3.83) & 68.11 (4.30) & 71.04 (4.19) & 71.67 (3.95) & 72.27 (4.21) & 73.79 (4.08) & 76.88 (4.31)\\

&MedBERT & 82.22 (3.85) & 70.57 (4.10) & 68.31 (3.77) & 65.86 (3.91) & 63.81 (4.12) & 64.61 (3.98) & 70.21 (4.06) & 73.39 (3.80) & 72.78 (3.93) & 74.36 (4.19) & 74.34 (4.11)& 79.00 (3.87) \\

&SciBERT & 84.58 (3.97) & 64.24 (3.49) & 64.01 (3.99)& 63.57 (3.88) & 64.19 (3.69) & 56.88 (3.89) & 66.70 (4.02) &  69.57 (4.17)& 70.01 (4.02)& 71.32 (3.82) & 72.47 (4.33)& 77.89 (3.66)\\
\midrule

\multirow{3}{*}{\textbf{Oxacillin}} & BioClinicalBERT & 80.53 (4.63) & 49.28 (4.78) & 47.52 (4.56) & 57.65 (4.94) & 66.16 (4.65) & 64.80 (4.97) & 67.32 (4.72) & 67.62 (4.88) & 68.52 (4.69) & 69.46 (4.90) & 68.75 (4.53) & 74.49 (4.87) \\

&MedBERT & 79.72 (4.98) & 49.95 (4.81) & 54.56 (4.73) & 63.34 (5.02) & 64.83 (4.95) & 61.81 (4.87) & 65.34 (5.13) & 66.36 (4.75) & 68.88 (5.08) & 69.28 (4.93) & 68.76 (4.79) & 74.02 (4.11)\\
&SciBERT & 80.33 (4.69) & 52.03 (4.77) & 50.40 (4.43) & 58.44 (4.69) & 61.47 (4.54) & 58.45 (4.91) & 65.03 (4.80) & 66.74 (4.64) & 68.39 (4.44) & 68.88 (4.38) & 68.62 (4.55) & 74.55 (4.20) \\

\midrule
\multirow{3}{*}{\textbf{Tetracycline}} & BioClinicalBERT & 86.28 (4.03) & 80.36 (4.45) & 81.96 (3.72) & 80.00 (4.38) & 78.80 (4.01) & 82.53 (3.75) & 81.46 (4.22) & 80.85 (3.98) & 81.75 (4.12) & 81.68 (3.94) & 81.08 (4.27) & 84.97 (3.88) \\
&MedBERT & 84.79 (4.32) & 80.84 (4.13) & 81.59 (4.1) & 79.07 (3.84) & 81.54 (4.11) & 79.83 (3.96) & 78.81 (3.91) & 79.39 (4.19) & 81.14 (3.9) & 81.82 (3.95) & 80.94 (3.88) & 81.13 (4.09)\\
&SciBERT & 87.64 (3.34) & 81.18 (3.77) & 80.83 (4.14) & 79.86 (3.99) & 78.62 (3.73) & 81.77 (3.86) & 82.73 (3.88) & 82.96 (3.23) & 82.36 (4.00) & 81.97 (3.67) & 82.31 (3.76) & 84.99 (3.77) \\
\midrule

\multirow{3}{*}{\textbf{Trimethoprim/sulfa}} & BioClinicalBERT & 87.76 (3.59) & 84.34 (3.14) & 84.31 (2.78) & 81.66 (3.23) & 80.91 (2.95) & 82.45 (3.10) & 81.48 (2.87) & 81.47 (3.35) & 80.47 (2.72) & 84.54 (2.63) & 84.77 (3.21) & 87.79 (2.84)
 \\
&MedBERT & 89.83 (3.12) & 83.14 (2.87) & 82.77 (2.99) & 78.61 (3.01) & 80.02 (3.18) & 83.66 (3.03) & 79.32 (2.92) & 81.19 (3.17) & 83.30 (2.85) & 83.47 (3.14) & 84.85 (2.97) & 86.77 (3.08)\\
&SciBERT & 90.04 (3.29) & 78.60 (2.77) & 79.19 (3.00) & 78.93 (3.13) & 78.92 (2.88) & 79.74 (2.75) & 77.33 (3.10) & 80.75 (3.52) & 84.26 (3.44) & 83.85 (2.39) & 84.34 (3.04) & 87.76 (3.14) \\
\midrule
\multirow{3}{*}{\textbf{Vancomycin}} & BioClinicalBERT & 78.81 (4.31) & 46.29 (4.02) & 45.78 (3.95) & 53.70 (4.08) & 61.58 (3.97) & 50.31 (4.10) & 49.99 (3.92) & 64.24 (4.07) & 63.25 (3.99) & 63.30 (4.01) & 63.43 (3.96) & 72.32 (4.05) \\
&MedBERT & 79.10 (4.05) & 43.88 (4.12) & 43.43 (3.95) & 47.30 (4.01) & 45.78 (3.89) & 46.39 (4.17) & 48.01 (3.85) & 56.62 (4.19) & 60.41 (3.96) & 63.71 (4.03) & 63.08 (3.92) & 70.99 (4.09) \\
&SciBERT & 80.15 (4.03) & 50.09 (3.97) & 46.34 (4.21) & 52.04 (3.76) & 53.26 (4.12) & 51.80 (3.89) & 52.99 (4.07) & 56.00 (3.98) & 61.15 (4.01) & 66.48 (4.28) & 64.69 (3.63) & 71.62 (3.85) \\

\bottomrule

\end{tabular}
\end{adjustbox}
\end{table}

\begin{table}[h]
\centering
\caption{Corresponding $\Delta$ FEET Table describing the AUROC Score Differentials on Various Antibiotic Tasks}
\begin{adjustbox}{width=1\textwidth}
\label{tab:Delta FEET Table}
\begin{tabular}{llcccccccccccc}
\toprule
\textbf{AUROC/Task} & \textbf{Models} & \textbf{Frozen} & \textbf{2 shot} & \textbf{4 shot} & \textbf{8 shot} & \textbf{16 shot} & \textbf{32 shot} & \textbf{64 shot} & \textbf{128 shot} & \textbf{256 shot} & \textbf{512 shot} & \textbf{1024 shot} & \textbf{Fine-tuned} \\
\midrule
\multirow{3}{*}{\textbf{Clindamycin}} & BioClinicalBERT & ------- &  -21.11\% &  -20.30\% &  -18.26\% &   -21.12\% &   -18.32\% &   -18.15\% &    -18.11\% &    -17.87\% &    -16.27\% &     -15.44\% &       -7.40\% \\
 & MedBERT&  ------ &  -19.94\% &  -18.73\% &  -18.83\% &   -19.87\% &   -17.98\% &   -16.95\% &    -16.66\% &    -15.90\% &    -14.94\% &     -13.36\% &       -4.87\% \\
  & SciBERT& ------ &  -19.80\% &  -22.38\% &  -21.21\% &   -23.87\% &   -21.80\% &   -19.88\% &    -17.57\% &    -17.06\% &    -17.38\% &     -15.10\% &       -5.67\% \\
  \midrule
  \multirow{3}{*}{\textbf{Erythromycin}} & BioClinicalBERT &------ &  -17.47\% &  -16.67\% &  -14.23\% &   -14.79\% &   -17.15\% &   -14.68\% &    -14.39\% &    -15.35\% &    -14.82\% &     -13.60\% &       -5.26\% \\
  &MedBERT&------ &  -16.13\% &  -18.91\% &  -16.07\% &   -15.85\% &   -19.59\% &   -19.58\% &    -15.36\% &    -13.87\% &    -12.81\% &     -11.81\% &       -3.73\% \\
  &SciBERT&------ &  -19.73\% &  -21.68\% &  -17.70\% &   -20.43\% &   -18.30\% &   -17.14\% &    -15.08\% &    -14.58\% &    -14.36\% &     -15.06\% &       -5.27\% \\
  \midrule
  \multirow{3}{*}{\textbf{Gentamicin}} & BioClinicalBERT &------ &  -15.58\% &  -17.56\% &  -15.46\% &   -16.26\% &   -21.31\% &   -17.65\% &    -18.87\% &    -19.84\% &    -15.74\% &      -8.79\% &       -4.40\% \\
  &MedBERT&------ &  -14.48\% &  -13.17\% &  -33.25\% &   -31.75\% &   -20.07\% &   -35.72\% &    -33.23\% &    -23.85\% &    -17.51\% &     -13.10\% &       -6.08\% \\
  &SciBERT&------ &  -28.45\% &  -22.04\% &  -25.74\% &   -25.23\% &   -22.63\% &   -25.28\% &    -20.80\% &    -21.56\% &    -28.78\% &     -22.52\% &      -15.75\% \\
  \midrule
  \multirow{3}{*}{\textbf{Levofloxacin}} & BioClinicalBERT &------ &  -20.81\% &  -20.69\% &  -19.15\% &   -22.57\% &   -24.72\% &   -21.24\% &    -16.55\% &    -14.78\% &    -13.59\% &     -12.79\% &      -10.85\% \\
  &MedBERT&------ &  -17.88\% &  -17.48\% &  -22.25\% &   -25.95\% &   -23.05\% &   -17.44\% &    -13.19\% &    -11.89\% &    -11.07\% &     -10.02\% &       -8.44\% \\
  &SciBERT&------ &  -26.24\% &  -28.11\% &  -27.68\% &   -26.04\% &   -37.00\% &   -21.80\% &    -16.80\% &    -15.99\% &    -15.24\% &     -13.85\% &      -12.13\% \\
  \midrule
  \multirow{3}{*}{\textbf{Oxacillin}} & BioClinicalBERT & ------ &  -41.73\% &  -42.48\% &  -28.40\% &   -15.42\% &   -18.99\% &   -13.99\% &    -15.08\% &    -12.94\% &    -13.03\% &     -12.01\% &      -10.94\% \\
  &MedBERT&------ &  -39.11\% &  -36.24\% &  -22.48\% &   -20.31\% &   -22.31\% &   -17.78\% &    -13.77\% &    -13.75\% &    -13.31\% &     -12.48\% &      -11.39\% \\
  &SciBERT&------ &  -36.82\% &  -39.96\% &  -28.20\% &   -25.12\% &   -29.43\% &   -18.51\% &    -15.69\% &    -15.50\% &    -13.97\% &     -13.21\% &      -11.36\% \\
  \midrule
  \multirow{3}{*}{\textbf{Tetracycline}} & BioClinicalBERT &------ &  -14.30\% &  -16.13\% &  -19.09\% &   -19.64\% &   -16.17\% &   -16.94\% &    -19.78\% &    -16.11\% &    -13.93\% &     -14.84\% &      -12.59\% \\
  &MedBERT&------ &   -9.74\% &  -12.59\% &  -19.60\% &   -14.72\% &   -19.52\% &   -18.60\% &    -17.56\% &    -14.02\% &    -12.10\% &      -7.95\% &       -7.07\% \\
  &SciBERT&------ &  -16.34\% &  -16.79\% &  -18.26\% &   -20.30\% &   -15.46\% &   -13.19\% &    -13.34\% &    -12.75\% &    -14.50\% &     -13.01\% &      -12.00\% \\
  \midrule
  \multirow{3}{*}{\textbf{Trimethoprim/sulfa}} & BioClinicalBERT &------ &  -14.58\% &  -15.32\% &  -20.65\% &   -23.08\% &   -20.36\% &   -21.49\% &    -22.93\% &    -17.55\% &    -13.95\% &     -10.43\% &       -8.54\% \\
  &MedBERT&------ &  -16.06\% &  -17.17\% &  -25.95\% &   -25.72\% &   -17.63\% &   -23.49\% &    -20.11\% &    -17.89\% &    -15.07\% &     -12.48\% &      -10.50\% \\
  &SciBERT&------ &  -27.63\% &  -26.60\% &  -25.78\% &   -26.97\% &   -23.56\% &   -32.49\% &    -27.35\% &    -19.73\% &    -16.88\% &     -16.25\% &      -14.67\% \\
  \midrule
  \multirow{3}{*}{\textbf{Vancomycin}} & BioClinicalBERT & ------ &  -36.12\% &  -37.07\% &  -23.69\% &   -17.07\% &   -29.73\% &   -28.13\% &    -14.50\% &    -14.01\% &    -13.71\% &     -13.30\% &      -11.37\% \\
  &MedBERT&------ &  -38.26\% &  -38.19\% &  -31.33\% &   -30.83\% &   -32.06\% &   -30.03\% &    -18.14\% &    -14.72\% &    -12.15\% &     -11.35\% &       -9.83\% \\
  &SciBERT&------ &  -30.58\% &  -35.87\% &  -26.82\% &   -23.26\% &   -25.89\% &   -22.43\% &    -20.34\% &    -17.12\% &    -13.87\% &     -14.10\% &      -11.76\% \\
\bottomrule

\textbf{AUPRC/Task} &  &  &  &  &  &  &  &  &  & &  &  & \\
\midrule
\multirow{3}{*}{\textbf{Clindamycin}} 
&BioClinicalBERT &------ & -29.63\% & -31.27\% & -28.42\% & -25.33\% & -23.49\% & -21.11\% & -18.37\% & -17.03\% & -14.90\% & -12.81\% & -8.74\% \\
&MedBERT &------ & -28.89\% & -30.47\% & -27.63\% & -25.15\% & -23.29\% & -21.00\% & -18.12\% & -15.96\% & -14.15\% & -11.90\% & -7.89\% \\
&SciBERT &------ & -28.68\% & -30.37\% & -27.50\% & -24.88\% & -23.02\% & -20.57\% & -17.90\% & -16.07\% & -14.35\% & -12.00\% & -7.77\% \\
\midrule
\multirow{3}{*}{\textbf{Erythromycin}} & BioClinicalBERT &------ & -21.81\% & -21.64\% & -19.96\% & -20.08\% & -20.12\% & -20.42\% & -20.42\% & -22.22\% & -22.03\% & -22.15\% & -15.27\% \\
&MedBERT &------ & -24.45\% & -25.34\% & -21.47\% & -22.14\% & -23.05\% & -24.41\% & -23.11\% & -22.09\% & -21.78\% & -21.94\% & -14.14\% \\
&SciBERT &------ & -23.87\% & -23.55\% & -23.24\% & -23.73\% & -23.01\% & -20.94\% & -22.86\% & -23.07\% & -21.88\% & -22.35\% & -13.60\% \\
\midrule
\multirow{3}{*}{\textbf{Gentamicin}} & BioClinicalBERT& ------ & -1.41\% & -1.43\% & -1.80\% & -1.93\% & -1.75\% & -1.65\% & -1.64\% & -2.13\% & -1.97\% & -1.66\% & -0.19\% \\
&MedBERT &------ & -1.06\% & -0.83\% & -2.95\% & -2.28\% & -1.65\% & -1.65\% & -5.14\% & -3.54\% & -2.28\% & -2.60\% & -1.31\% \\
&SciBERT &------ & -1.99\% & -1.16\% & -1.47\% & -1.07\% & -1.07\% & -1.12\% & -0.89\% & -2.24\% & -2.15\% & -1.53\% & -0.67\% \\
\midrule
\multirow{3}{*}{\textbf{Levofloxacin}} & BioClinicalBERT &------ & -14.84\% & -16.37\% & -16.10\% & -18.19\% & -19.54\% & -16.30\% & -13.37\% & -12.74\% & -12.14\% & -10.62\% & -7.53\% \\
&MedBERT &------ & -11.65\% & -13.91\% & -16.36\% & -18.41\% & -17.61\% & -12.01\% & -8.83\% & -9.44\% & -7.86\% & -7.88\% & -3.22\% \\
&SciBERT &------ & -20.34\% & -20.57\% & -21.01\% & -20.39\% & -27.70\% & -17.88\% & -15.01\% & -14.57\% & -13.26\% & -12.11\% & -6.69\% \\
\midrule
\multirow{3}{*}{\textbf{Oxacillin}} & BioClinicalBERT & ------ & -31.25\% & -33.01\% & -22.88\% & -14.37\% & -15.73\% & -13.21\% & -12.91\% & -12.01\% & -11.07\% & -11.78\% & -6.04\% \\
&MedBERT &------ & -29.77\% & -25.16\% & -16.38\% & -14.89\% & -17.91\% & -14.38\% & -13.36\% & -10.84\% & -10.44\% & -10.96\% & -5.70\% \\
&SciBERT &------ & -28.30\% & -29.93\% & -21.89\% & -18.86\% & -21.88\% & -15.30\% & -13.59\% & -11.94\% & -11.45\% & -11.71\% & -5.78\% \\
\midrule
\multirow{3}{*}{\textbf{Tetracycline}} & BioClinicalBERT &------ & -5.92\% & -4.32\% & -6.28\% & -7.48\% & -3.75\% & -4.82\% & -5.43\% & -4.53\% & -4.60\% & -5.20\% & -1.31\% \\
&MedBERT &------ & -3.95\% & -3.20\% & -5.72\% & -3.25\% & -4.96\% & -5.98\% & -5.40\% & -3.65\% & -2.97\% & -3.85\% & -3.66\% \\
&SciBERT &------ & -6.46\% & -6.81\% & -7.78\% & -9.02\% & -5.87\% & -4.91\% & -4.68\% & -5.28\% & -5.67\% & -5.33\% & -2.65\% \\
\midrule
\multirow{3}{*}{\textbf{Trimethoprim/sulfa}} & BioClinicalBERT &  ------ & -3.42\% & -3.45\% & -6.10\% & -6.85\% & -5.31\% & -6.28\% & -6.29\% & -7.29\% & -3.22\% & -2.99\% & 0.03\% \\
&MedBERT &------ & -6.69\% & -7.06\% & -11.22\% & -9.81\% & -6.17\% & -10.51\% & -8.64\% & -6.53\% & -6.36\% & -4.98\% & -3.06\% \\
&SciBERT &------ & -11.44\% & -10.85\% & -11.11\% & -11.12\% & -10.30\% & -12.71\% & -9.29\% & -5.78\% & -6.19\% & -5.70\% & -2.28\% \\
\midrule
\multirow{3}{*}{\textbf{Vancomycin}} & BioClinicalBERT &  ------ &  -32.52\% &  -33.03\% &  -25.11\% &   -17.23\% &   -28.50\% &   -28.82\% &    -14.57\% &    -15.56\% &    -15.51\% &     -15.38\% &       -6.49\% \\
&MedBERT &  ------ &  -35.22\% &  -35.67\% &  -31.80\% &   -33.32\% &   -32.71\% &   -31.09\% &    -22.48\% &    -18.69\% &    -15.39\% &     -16.02\% &       -8.11\% \\
&SciBERT &  ------ & -30.06\% & -33.81\% & -28.11\% & -26.89\% & -28.35\% & -27.16\% & -24.15\% & -19.00\% & -13.67\% & -15.46\% & -8.53\% \\
\bottomrule

\end{tabular}
\end{adjustbox}
\end{table}

The FEET table (Table 5) displays the AUROC scores for various biomedical foundation models—BioClinicalBERT, MedBERT, and SciBERT—across several antibiotic prediction tasks.  Across all tasks, there is a clear performance trend where models generally degrade as they move from Frozen to Fine-tuned settings. For instance, in the Clindamycin task, BioClinicalBERT shows an AUROC decreases from 74.99 in the Frozen setting to 67.59 after full fine-tuning, despite some fluctuation in the intermediate shot settings. MedBERT and SciBERT follow a similar pattern, though MedBERT consistently outperforms SciBERT across most tasks.

The \(\Delta\) FEET Table (Table 6) highlights the relative performance changes across embedding types, using the Frozen setting as a baseline. The table indicates that models tend to show a considerable performance drop in the Few-shot settings compared to the Frozen baseline, especially in the early shots (2, 4, and 8), with recovery occurring as more shots are added or as full fine-tuning is applied. 

\section{Discussion}
\subsection{A Principled Approach to Foundation Model Research}

In our study, the FEET Tables demonstrate their essential role in benchmarking foundation models. These tables summarize key performance metrics and aid decision-making by showing which models (e.g., frozen versus fine-tuned) are best suited for specific tasks. By following a standardized protocol, FEET Tables help ensure that model selection aligns with study objectives, enhancing the reliability and reproducibility of machine learning applications. This approach fosters a more rigorous and informed research environment while eliminating inconsistencies and arbitrary benchmarks.

In our Sentiment Analysis case study, we observed a seemingly trivial result: when we train our foundation models and adjust the model parameters from few-shot to fully fine-tuned, there is an expected increase in performance. While this may appear obvious, the study also revealed that BERT, DistilBERT, and GPT2 were not very effective sentiment analysis tools in their frozen embedding settings, providing insights into the limited transferability of their pre-training for this specific task. While we expect this pattern to hold in most fine-tuning studies, it is important to note that this is not always the case.

\subsection{Results reveals that fine-tuning can degrade performance in the Medical Case Study}

Our analysis reveals a counterintuitive finding: fine-tuning can sometimes degrade model performance. This insight was previously found but can happen for many reasons \citep{mukhoti2023fine}. This may result from overfitting on small, unrepresentative datasets, suboptimal hyperparameter settings, or the distortion of pre-trained representations due to new data perturbations \citep{kumar2022fine}. We also observe that different foundation models perform optimally under different conditions (Frozen embeddings favor Bio\_ClinicalBERT, while Few-shot and Fine-tuned embeddings benefit MedBERT). This type of reporting not only highlights these critical differences but also emphasizes the importance of selecting the right model for each specific task, as evidenced by the varying \(\Delta\) values across models. These insights formalize and guide us toward more targeted and effective model selection in application studies, underscoring the need for robust, task-specific model evaluation frameworks.

\section{Conclusion}

In this work, we introduced FEET, a protocol designed to standardize the evaluation of foundation models across different embedding types—Frozen, Few-shot, and Fine-tuned. By implementing FEET, researchers and practitioners can gain detailed insights into the relative performance improvements (or declines), denoted by \( \Delta \), that each embedding technique offers compared to the baseline performance of frozen embeddings. This structured approach not only deepens the understanding of each model's capabilities but also helps in identifying which models are most suitable for specific tasks. 

\paragraph{Future Work:} Future work includes releasing a codebase that easily applies this framework to models from HuggingFace. We also believe the commentary from the community could provide provide helpful inisghts as to how to interpret results under different conditions. We are encouraged by our preliminary design and if this project is something you would like to implement and build, feel free to reach out to the authors.

\paragraph*{Data and Code Availability}
Code to replicate our work can be found in \footnote{\href{https://github.com/Simonlee711/FEET}{https://github.com/Simonlee711/FEET}}.
Our work uses the MIMIC dataset which can be accessed after undergoing an approval process. Meanwhile, SST-2 is publicly available for download.

{
\small
\bibliographystyle{plainnat}
\bibliography{Styles/neurips_2024}

\begin{thebibliography}{41}
\providecommand{\natexlab}[1]{#1}
\providecommand{\url}[1]{\texttt{#1}}
\expandafter\ifx\csname urlstyle\endcsname\relax
  \providecommand{\doi}[1]{doi: #1}\else
  \providecommand{\doi}{doi: \begingroup \urlstyle{rm}\Url}\fi

\bibitem[Alsentzer et~al.(2019)Alsentzer, Murphy, Boag, Weng, Jin, Naumann, and McDermott]{alsentzer2019publicly}
Emily Alsentzer, John~R Murphy, Willie Boag, Wei-Hung Weng, Di~Jin, Tristan Naumann, and Matthew McDermott.
\newblock Publicly available clinical bert embeddings.
\newblock \emph{arXiv preprint arXiv:1904.03323}, 2019.

\bibitem[Beltagy et~al.(2019)Beltagy, Lo, and Cohan]{beltagy2019scibert}
Iz~Beltagy, Kyle Lo, and Arman Cohan.
\newblock Scibert: A pretrained language model for scientific text.
\newblock \emph{arXiv preprint arXiv:1903.10676}, 2019.

\bibitem[Birk et~al.(2024)Birk, Hallin, and Kasieczka]{birk2024omnijet}
Joschka Birk, Anna Hallin, and Gregor Kasieczka.
\newblock Omnijet-$\alpha$: the first cross-task foundation model for particle physics.
\newblock \emph{Machine Learning: Science and Technology}, 2024.

\bibitem[Bommasani et~al.(2021)Bommasani, Hudson, Adeli, Altman, Arora, von Arx, Bernstein, Bohg, Bosselut, Brunskill, et~al.]{bommasani2021opportunities}
Rishi Bommasani, Drew~A Hudson, Ehsan Adeli, Russ Altman, Simran Arora, Sydney von Arx, Michael~S Bernstein, Jeannette Bohg, Antoine Bosselut, Emma Brunskill, et~al.
\newblock On the opportunities and risks of foundation models.
\newblock \emph{arXiv preprint arXiv:2108.07258}, 2021.

\bibitem[Brown(2020)]{brown2020language}
Tom~B Brown.
\newblock Language models are few-shot learners.
\newblock \emph{arXiv preprint ArXiv:2005.14165}, 2020.

\bibitem[Church(2017)]{church2017word2vec}
Kenneth~Ward Church.
\newblock Word2vec.
\newblock \emph{Natural Language Engineering}, 23\penalty0 (1):\penalty0 155--162, 2017.

\bibitem[Cobbe et~al.(2021)Cobbe, Kosaraju, Bavarian, Chen, Jun, Kaiser, Plappert, Tworek, Hilton, Nakano, et~al.]{cobbe2021training}
Karl Cobbe, Vineet Kosaraju, Mohammad Bavarian, Mark Chen, Heewoo Jun, Lukasz Kaiser, Matthias Plappert, Jerry Tworek, Jacob Hilton, Reiichiro Nakano, et~al.
\newblock Training verifiers to solve math word problems.
\newblock \emph{arXiv preprint arXiv:2110.14168}, 2021.

\bibitem[Devlin(2018)]{devlin2018bert}
Jacob Devlin.
\newblock Bert: Pre-training of deep bidirectional transformers for language understanding.
\newblock \emph{arXiv preprint arXiv:1810.04805}, 2018.

\bibitem[Ding et~al.(2023)Ding, Qin, Yang, Wei, Yang, Su, Hu, Chen, Chan, Chen, et~al.]{ding2023parameter}
Ning Ding, Yujia Qin, Guang Yang, Fuchao Wei, Zonghan Yang, Yusheng Su, Shengding Hu, Yulin Chen, Chi-Min Chan, Weize Chen, et~al.
\newblock Parameter-efficient fine-tuning of large-scale pre-trained language models.
\newblock \emph{Nature Machine Intelligence}, 5\penalty0 (3):\penalty0 220--235, 2023.

\bibitem[Dodge et~al.(2020)Dodge, Ilharco, Schwartz, Farhadi, Hajishirzi, and Smith]{dodge2020fine}
Jesse Dodge, Gabriel Ilharco, Roy Schwartz, Ali Farhadi, Hannaneh Hajishirzi, and Noah Smith.
\newblock Fine-tuning pretrained language models: Weight initializations, data orders, and early stopping.
\newblock \emph{arXiv preprint arXiv:2002.06305}, 2020.

\bibitem[Ericsson et~al.(2022)Ericsson, Gouk, Loy, and Hospedales]{ericsson2022self}
Linus Ericsson, Henry Gouk, Chen~Change Loy, and Timothy~M Hospedales.
\newblock Self-supervised representation learning: Introduction, advances, and challenges.
\newblock \emph{IEEE Signal Processing Magazine}, 39\penalty0 (3):\penalty0 42--62, 2022.

\bibitem[Hendrycks et~al.(2020)Hendrycks, Burns, Basart, Zou, Mazeika, Song, and Steinhardt]{hendrycks2020measuring}
Dan Hendrycks, Collin Burns, Steven Basart, Andy Zou, Mantas Mazeika, Dawn Song, and Jacob Steinhardt.
\newblock Measuring massive multitask language understanding.
\newblock \emph{arXiv preprint arXiv:2009.03300}, 2020.

\bibitem[Hu et~al.(2021)Hu, Shen, Wallis, Allen-Zhu, Li, Wang, Wang, and Chen]{hu2021lora}
Edward~J Hu, Yelong Shen, Phillip Wallis, Zeyuan Allen-Zhu, Yuanzhi Li, Shean Wang, Lu~Wang, and Weizhu Chen.
\newblock Lora: Low-rank adaptation of large language models.
\newblock \emph{arXiv preprint arXiv:2106.09685}, 2021.

\bibitem[Jiang et~al.(2024)Jiang, Ge, Ge, Shi, Yuan, and Shan]{jiang2024supervised}
Xiaohu Jiang, Yixiao Ge, Yuying Ge, Dachuan Shi, Chun Yuan, and Ying Shan.
\newblock Supervised fine-tuning in turn improves visual foundation models.
\newblock \emph{arXiv preprint arXiv:2401.10222}, 2024.

\bibitem[Johnson et~al.(2020)Johnson, Bulgarelli, Pollard, Horng, Celi, and Mark]{johnson2020mimic}
Alistair Johnson, Lucas Bulgarelli, Tom Pollard, Steven Horng, Leo~Anthony Celi, and Roger Mark.
\newblock Mimic-iv.
\newblock \emph{PhysioNet. Available online at: https://physionet. org/content/mimiciv/1.0/(accessed August 23, 2021)}, pages 49--55, 2020.

\bibitem[Joulin et~al.(2016)Joulin, Grave, Bojanowski, Douze, J{\'e}gou, and Mikolov]{joulin2016fasttext}
Armand Joulin, Edouard Grave, Piotr Bojanowski, Matthijs Douze, H{\'e}rve J{\'e}gou, and Tomas Mikolov.
\newblock Fasttext. zip: Compressing text classification models.
\newblock \emph{arXiv preprint arXiv:1612.03651}, 2016.

\bibitem[Karimi~Mahabadi et~al.(2021)Karimi~Mahabadi, Henderson, and Ruder]{karimi2021compacter}
Rabeeh Karimi~Mahabadi, James Henderson, and Sebastian Ruder.
\newblock Compacter: Efficient low-rank hypercomplex adapter layers.
\newblock \emph{Advances in Neural Information Processing Systems}, 34:\penalty0 1022--1035, 2021.

\bibitem[Kumar et~al.(2022)Kumar, Raghunathan, Jones, Ma, and Liang]{kumar2022fine}
Ananya Kumar, Aditi Raghunathan, Robbie Jones, Tengyu Ma, and Percy Liang.
\newblock Fine-tuning can distort pretrained features and underperform out-of-distribution.
\newblock \emph{arXiv preprint arXiv:2202.10054}, 2022.

\bibitem[Lai et~al.(2020)Lai, Dernoncourt, and Nguyen]{lai2020extensively}
Viet~Dac Lai, Franck Dernoncourt, and Thien~Huu Nguyen.
\newblock Extensively matching for few-shot learning event detection.
\newblock \emph{arXiv preprint arXiv:2006.10093}, 2020.

\bibitem[Lanusse et~al.(2023)Lanusse, Parker, Golkar, Cranmer, Bietti, Eickenberg, Krawezik, McCabe, Ohana, Pettee, et~al.]{lanusse2023astroclip}
Francois Lanusse, Liam Parker, Siavash Golkar, Miles Cranmer, Alberto Bietti, Michael Eickenberg, Geraud Krawezik, Michael McCabe, Ruben Ohana, Mariel Pettee, et~al.
\newblock Astroclip: Cross-modal pre-training for astronomical foundation models.
\newblock \emph{arXiv preprint arXiv:2310.03024}, 2023.

\bibitem[Lee et~al.(2024{\natexlab{a}})Lee, Brokowski, and Chiang]{lee2024enhancing}
Simon~A Lee, Trevor Brokowski, and Jeffrey~N Chiang.
\newblock Enhancing antibiotic stewardship using a natural language approach for better feature representation.
\newblock \emph{arXiv preprint arXiv:2405.20419}, 2024{\natexlab{a}}.

\bibitem[Lee et~al.(2024{\natexlab{b}})Lee, Jain, Chen, Biswas, Fang, Rudas, and Chiang]{lee2024multimodal}
Simon~A Lee, Sujay Jain, Alex Chen, Arabdha Biswas, Jennifer Fang, Akos Rudas, and Jeffrey~N Chiang.
\newblock Multimodal clinical pseudo-notes for emergency department prediction tasks using multiple embedding model for ehr (meme).
\newblock \emph{arXiv preprint arXiv:2402.00160}, 2024{\natexlab{b}}.

\bibitem[Majdik et~al.(2024)Majdik, Graham, Shiva~Edward, Rodriguez, Karnes, Jensen, Barbour, and Rousseau]{majdik2024sample}
Zoltan~P Majdik, S~Scott Graham, Jade~C Shiva~Edward, Sabrina~N Rodriguez, Martha~S Karnes, Jared~T Jensen, Joshua~B Barbour, and Justin~F Rousseau.
\newblock Sample size considerations for fine-tuning large language models for named entity recognition tasks: Methodological study.
\newblock \emph{JMIR AI}, 3:\penalty0 e52095, 2024.

\bibitem[Mukhoti et~al.(2023)Mukhoti, Gal, Torr, and Dokania]{mukhoti2023fine}
Jishnu Mukhoti, Yarin Gal, Philip~HS Torr, and Puneet~K Dokania.
\newblock Fine-tuning can cripple your foundation model; preserving features may be the solution.
\newblock \emph{arXiv preprint arXiv:2308.13320}, 2023.

\bibitem[Ono and Lee(2024)]{ono2024text}
Kyoka Ono and Simon~A Lee.
\newblock Text serialization and their relationship with the conventional paradigms of tabular machine learning.
\newblock \emph{arXiv preprint arXiv:2406.13846}, 2024.

\bibitem[Parnami and Lee(2022)]{parnami2022learning}
Archit Parnami and Minwoo Lee.
\newblock Learning from few examples: A summary of approaches to few-shot learning.
\newblock \emph{arXiv preprint arXiv:2203.04291}, 2022.

\bibitem[Pennington et~al.(2014)Pennington, Socher, and Manning]{pennington2014glove}
Jeffrey Pennington, Richard Socher, and Christopher~D Manning.
\newblock Glove: Global vectors for word representation.
\newblock In \emph{Proceedings of the 2014 conference on empirical methods in natural language processing (EMNLP)}, pages 1532--1543, 2014.

\bibitem[Radford(2018)]{radford2018improving}
Alec Radford.
\newblock Improving language understanding by generative pre-training.
\newblock 2018.

\bibitem[Radford et~al.(2019)Radford, Wu, Child, Luan, Amodei, Sutskever, et~al.]{radford2019language}
Alec Radford, Jeffrey Wu, Rewon Child, David Luan, Dario Amodei, Ilya Sutskever, et~al.
\newblock Language models are unsupervised multitask learners.
\newblock \emph{OpenAI blog}, 1\penalty0 (8):\penalty0 9, 2019.

\bibitem[Radford et~al.(2021)Radford, Kim, Hallacy, Ramesh, Goh, Agarwal, Sastry, Askell, Mishkin, Clark, et~al.]{radford2021learning}
Alec Radford, Jong~Wook Kim, Chris Hallacy, Aditya Ramesh, Gabriel Goh, Sandhini Agarwal, Girish Sastry, Amanda Askell, Pamela Mishkin, Jack Clark, et~al.
\newblock Learning transferable visual models from natural language supervision.
\newblock In \emph{International conference on machine learning}, pages 8748--8763. PMLR, 2021.

\bibitem[Sanh(2019)]{sanh2019distilbert}
V~Sanh.
\newblock Distilbert, a distilled version of bert: Smaller, faster, cheaper and lighter.
\newblock \emph{arXiv preprint arXiv:1910.01108}, 2019.

\bibitem[Socher et~al.(2013)Socher, Perelygin, Wu, Chuang, Manning, Ng, and Potts]{socher2013recursive}
Richard Socher, Alex Perelygin, Jean Wu, Jason Chuang, Christopher~D Manning, Andrew~Y Ng, and Christopher Potts.
\newblock Recursive deep models for semantic compositionality over a sentiment treebank.
\newblock In \emph{Proceedings of the 2013 conference on empirical methods in natural language processing}, pages 1631--1642, 2013.

\bibitem[Sung et~al.(2022)Sung, Cho, and Bansal]{sung2022vl}
Yi-Lin Sung, Jaemin Cho, and Mohit Bansal.
\newblock Vl-adapter: Parameter-efficient transfer learning for vision-and-language tasks.
\newblock In \emph{Proceedings of the IEEE/CVF conference on computer vision and pattern recognition}, pages 5227--5237, 2022.

\bibitem[Theis et~al.(2015)Theis, Oord, and Bethge]{theis2015note}
Lucas Theis, A{\"a}ron van~den Oord, and Matthias Bethge.
\newblock A note on the evaluation of generative models.
\newblock \emph{arXiv preprint arXiv:1511.01844}, 2015.

\bibitem[Torrey and Shavlik(2010)]{torrey2010transfer}
Lisa Torrey and Jude Shavlik.
\newblock Transfer learning.
\newblock In \emph{Handbook of research on machine learning applications and trends: algorithms, methods, and techniques}, pages 242--264. IGI global, 2010.

\bibitem[Vasantharajan et~al.(2022)Vasantharajan, Tun, Thi-Nga, Jain, Rong, and Siong]{vasantharajan2022medbert}
Charangan Vasantharajan, Kyaw~Zin Tun, Ho~Thi-Nga, Sparsh Jain, Tong Rong, and Chng~Eng Siong.
\newblock Medbert: a pre-trained language model for biomedical named entity recognition.
\newblock In \emph{2022 Asia-Pacific Signal and Information Processing Association Annual Summit and Conference (APSIPA ASC)}, pages 1482--1488. IEEE, 2022.

\bibitem[Vaswani(2017)]{vaswani2017attention}
Ashish Vaswani.
\newblock Attention is all you need.
\newblock \emph{arXiv preprint arXiv:1706.03762}, 2017.

\bibitem[Wang et~al.(2020)Wang, Yao, Kwok, and Ni]{wang2020generalizing}
Yaqing Wang, Quanming Yao, James~T Kwok, and Lionel~M Ni.
\newblock Generalizing from a few examples: A survey on few-shot learning.
\newblock \emph{ACM computing surveys (csur)}, 53\penalty0 (3):\penalty0 1--34, 2020.

\bibitem[Weiss et~al.(2016)Weiss, Khoshgoftaar, and Wang]{weiss2016survey}
Karl Weiss, Taghi~M Khoshgoftaar, and DingDing Wang.
\newblock A survey of transfer learning.
\newblock \emph{Journal of Big data}, 3:\penalty0 1--40, 2016.

\bibitem[Wolf et~al.(2019)Wolf, Debut, Sanh, Chaumond, Delangue, Moi, Cistac, Rault, Louf, Funtowicz, et~al.]{wolf2019huggingface}
Thomas Wolf, Lysandre Debut, Victor Sanh, Julien Chaumond, Clement Delangue, Anthony Moi, Pierric Cistac, Tim Rault, R{\'e}mi Louf, Morgan Funtowicz, et~al.
\newblock Huggingface's transformers: State-of-the-art natural language processing.
\newblock \emph{arXiv preprint arXiv:1910.03771}, 2019.

\bibitem[Wornow et~al.(2023)Wornow, Xu, Thapa, Patel, Steinberg, Fleming, Pfeffer, Fries, and Shah]{wornow2023shaky}
Michael Wornow, Yizhe Xu, Rahul Thapa, Birju Patel, Ethan Steinberg, Scott Fleming, Michael~A Pfeffer, Jason Fries, and Nigam~H Shah.
\newblock The shaky foundations of large language models and foundation models for electronic health records.
\newblock \emph{npj Digital Medicine}, 6\penalty0 (1):\penalty0 135, 2023.

\end{thebibliography}

%%%%%%%%%%%%%%%%%%%%%%%%%%%%%%%%%%%%%%%%%%%%%%%%%%%%%%%%%%%%

%%%%%%%%%%%%%%%%%%%%%%%%%%%%%%%%%%%%%%%%%%%%%%%%%%%%%%%%%%%%

% \appendix
% \onecolumn

% \newpage
% \section{Foundation Model Hyperparameters}

% We implement both stages utilizing the Huggingface transformers framework\footnote{\href{https://huggingface.co/docs/transformers/en/index}{https://huggingface.co/docs/transformers/en/index}}. The default AdamW optimizer, along with a linear learning rate scheduler, is used throughout all training phases. All foundation models are evaluated for 100 epochs in the fine-tuning with early stopping implemented at a learning rate of $2 \times 10^{-5}$, after which the model approaches convergence. During the finetuning phase, checkpoints are created every 50 minibatches. The model is then restored from the checkpoint exhibiting the lowest validation loss, monitored across a maximum of 10 epochs. The entire computational pipeline is executable on a single 48GB A100 GPU within a 24-hour timeframe.

% \section{Example of Arbitrary Shots in the Wild}
% \label{wild}

% \begin{figure}[h]
% \centering
% \includegraphics[width=0.9\textwidth]{benchmark.jpeg}
% \caption{A working example comparing the Claude Model against GPT and Gemini. Why are the number of shots arbitrary here? Why do they select 0? then 25? 4? 3? 10? It seems rather inconsistent and cherry-picked to showcase that Claude is the SOTA here.}
% \label{fig:teximage}
% \end{figure}

\end{document}